\documentclass[11pt]{article}

\usepackage{acl}

\usepackage{times}
\usepackage{latexsym}

\usepackage[T1]{fontenc}

\usepackage[utf8]{inputenc}

\usepackage{microtype}

\usepackage{inconsolata}

\usepackage{graphicx}

\usepackage{algorithm}
\usepackage{algorithmic}
\usepackage{url}            
\usepackage{booktabs}       
\usepackage{amsfonts}       
\usepackage{nicefrac}       
\usepackage{microtype}      
\usepackage{xcolor}         
\usepackage{xspace}
\usepackage{colortbl}
\usepackage{multirow}
\usepackage{graphicx}
\usepackage{caption}
\usepackage{amsmath}
\usepackage{subfig}
\usepackage{threeparttable}
\usepackage{bbding}  
\newcommand{\ModelName}{{AesTest}\xspace}

\title{AesTest: Measuring Aesthetic Intelligence from Perception to Production}

\author{
 \textbf{Guolong Wang\textsuperscript{1}},
 \textbf{Heng Huang\textsuperscript{2}},
 \textbf{Zhiqiang Zhang\textsuperscript{1}},
 \textbf{Wentian Li\textsuperscript{3}},
 \textbf{Feilong Ma\textsuperscript{3}},
 \textbf{Xin Jin\textsuperscript{4,5}}
\\
\\
 \textsuperscript{1}University of International Business and Economics,
\\
 \textsuperscript{2}University of Science and Technology of China,
\\
 \textsuperscript{3}Huawei Technologies Co., Ltd,
 \textsuperscript{4}Beijing Electronic Science and Technology Institute,
\\
 \textsuperscript{5}Beijing Institute for General Artificial Intelligence
\\
 \small{
   \textbf{Correspondence:} \href{mailto:jinxinbesti @foxmail.com}{jinxinbesti@foxmail.com}
 }
}

\begin{document}
\maketitle
\begin{abstract}
Perceiving and producing aesthetic judgments is a fundamental yet underexplored capability for multimodal large language models (MLLMs). However, existing benchmarks for image aesthetic assessment (IAA) are narrow in perception scope or lack the diversity needed to evaluate systematic aesthetic production. To address this gap, we introduce AesTest, a comprehensive benchmark for multimodal aesthetic perception and production, distinguished by the following features: 1) It consists of curated multiple-choice questions spanning ten tasks, covering perception, appreciation, creation, and photography. These tasks are grounded in psychological theories of generative learning. 2) It integrates data from diverse sources, including professional editing workflows, photographic composition tutorials, and crowdsourced preferences. It ensures coverage of both expert-level principles and real-world variation. 3) It supports various aesthetic query types, such as attribute-based analysis, emotional resonance, compositional choice, and stylistic reasoning. We evaluate both instruction-tuned IAA MLLMs and general MLLMs on AesTest, revealing significant challenges in building aesthetic intelligence.
We will publicly release AesTest to support future research in this area.
\end{abstract}

\begin{figure}
	\centering
	\includegraphics[width=\linewidth]{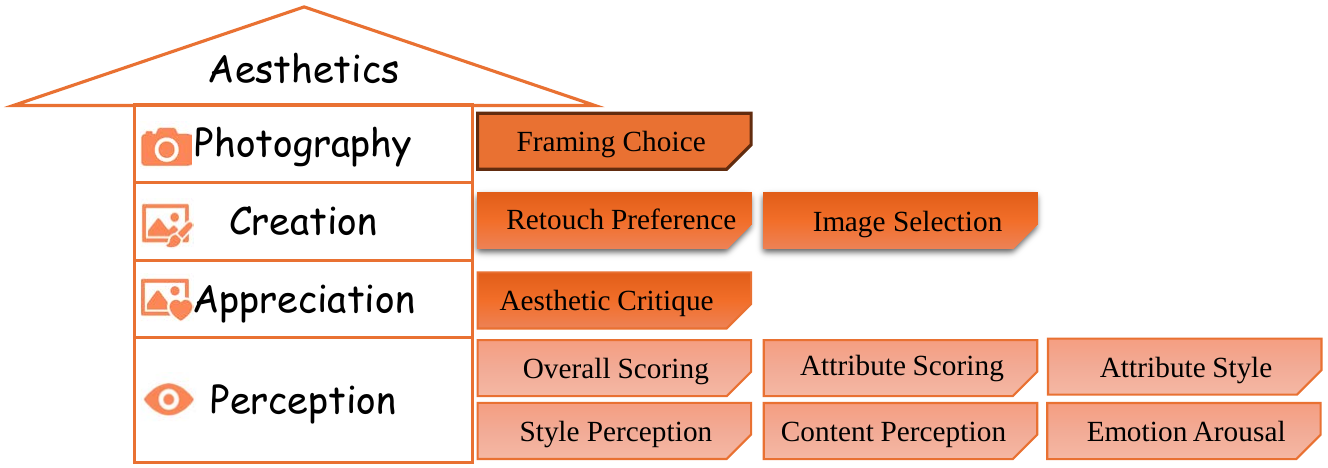} 
	\caption{The overall architecture of our \ModelName. It has four tasks: Perception, Appreciation, Creation, and Photography. Perception assesses image aesthetics (e.g., score, content, style, emotion). Appreciation elicits connoisseur-level critique. Creation tests a model’s ability to improve aesthetics via curation and retouching. Photography probes knowledge of principles for capturing impactful images.}
	\label{fig:overview}
\end{figure}

\section{Introduction}
Aesthetic experience arises from neurobiological processes, manifesting through both low- and high-level visual features~\cite{iigaya2021aesthetic}. Prior research has modeled human aesthetics via holistic perception~\cite{kao2016hierarchical,jin2018predicting,he2022rethinking,ke2023vila} or by decomposing specific aesthetic attributes~\cite{jin2019aesthetic,he2024rethinking,liu2024elta}. Recent efforts further explore individual and experiential differences~\cite{wang2018collaborative,li2020personality,nadal2025sensory}, aligning with the multimodal large language model (MLLM) paradigm for image understanding~\cite{deng2024enhancing}. Based on this, studies have begun to apply MLLMs for Image Aesthetic Assessment (IAA)~\cite{sheng2023aesclip,huang2024aesexpert,zhou2024uniaa,jiang2025multimodal}. However, there is currently a lack of unified testing standards, thus giving rise to two significant research challenges~\cite{huang2024aesbench}: 

1) \textit{What tasks can be designed to comprehensively interrogate aesthetic intelligence of large models?} While many works have explored specific subdomains of aesthetics, such as score regression or caption-based evaluation~\cite{ke2023vila,wang2023exploring,wang2024keep}, most lack a unified framework that integrates perceptual, appreciative, and productive dimensions. Current benchmarks remain fragmented, often limited to perception capability, not production capability~\cite{huang2024aesbench}.

2) \textit{How can diverse task modalities be unified to deliver quantifiable metrics?} Current tests contain tasks vary severely in format and scale, preventing direct comparison across capabilities. They often rely on crowd-sourced or professional labels without clarifying what kind of aesthetic intelligence they aim to probe. This ambiguity limits interpretability and generalization~\cite{dutt2025monetgpt}.

To address the above challenges, inspired by generative learning theories~\cite{fiorella2016eight}, which posit that learning is robust considering both understanding and production, we propose \ModelName, a new benchmark that comprehensively evaluates aesthetic intelligence through both understanding (perception \& appreciation) and production (creation \& photography), as shown in Figure~\ref{fig:overview}, introducing the following key features:

1) \textbf{Inclusion of diverse images}. Our \ModelName incorporates cross-domain image diversity as a fundamental design principle, ensuring comprehensive coverage across stylistic and contextual dimensions, including seven public datasets. 

2) \textbf{Rich tasks from aesthetic perception to production}. Our \ModelName has four tasks and ten sub-tasks, such as aesthetic critique and framing choice. This sets it apart from existing benchmarks focusing on perception only. Moreover, it covers classification, regression, and text generation tasks without relying on separate output modalities.   

3) \textbf{Unified multiple-choice question (MCQ) format}. 
Our \ModelName adopts a unified MCQ format, facilitating scalable and interpretable assessment across diverse tasks, to mitigate potential subjective interference during assessment. 

We conduct comprehensive experiments on \ModelName, where general and IAA MLLMs are studied. Our evaluation in \ModelName offers a comprehensive view of existing MLLMs' performance across diverse aesthetic tasks and input formats, revealing that aesthetic intelligence remains a challenging problem in both accuracy and generalization.

In summary, our contributions are threefold:

1) We introduce \ModelName, the first comprehensive, specialized benchmark for aesthetic intelligence, enabling evaluation of aesthetic competence from perception to production. 

2) \ModelName is constructed using diverse images and covers a wide range of aesthetic tasks. All tasks are re-organized from well-annotated public datasets and our carefully constructed items.

3) We perform extensive experiments based on \ModelName, whose results show the challenge of MLLMs in producing aesthetics.

\begin{figure}[tb!]
	\centering
	\includegraphics[width=\linewidth]{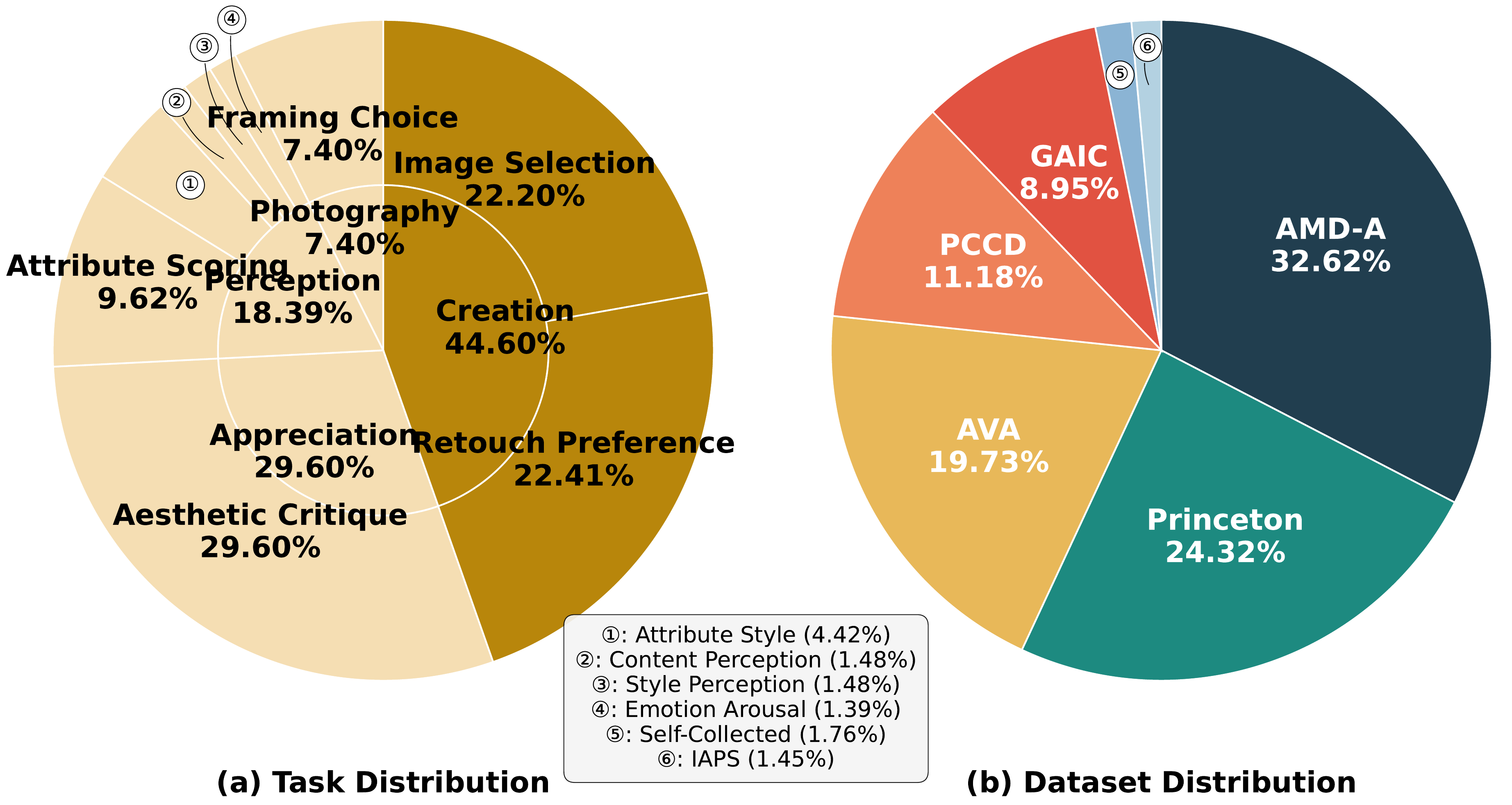} 
	\caption{Dataset statistics. (a) Task distribution, (b) Dataset distribution, where Self-Collected refers to images we curated specifically for \ModelName.}
	\label{fig:data}
\end{figure}

\begin{table}[tb!]
	\centering
            \resizebox{0.98\linewidth}{!}{
			\begin{threeparttable}
				\begin{tabular}{lccccc}
					\toprule
					\textbf{Benchmark} & Label & Production-included? & Task-oriented? & \#Images & \#QA\\
					\midrule
                    \textbf{AesBench}~\cite{huang2024aesbench} & Human & \XSolidBrush & \Checkmark & 2,800 & 8,400 \\
                    \textbf{UNIAA}~\cite{zhou2024uniaa} & Human & \XSolidBrush & \Checkmark & 5,855  & 5,354 \\
                    \midrule
                    \textbf{\ModelName} & Human & \Checkmark & \Checkmark & 17,885 & 8757 \\
				\bottomrule
				\end{tabular}			
				\end{threeparttable}}
        \caption{Comparison of popular IAA benchmarks and our proposed \ModelName benchmark.}
		\label{tab:benchcompare}	
	\end{table}  

\begin{figure*}[tb!]
	\centering
	\includegraphics[width=\linewidth]{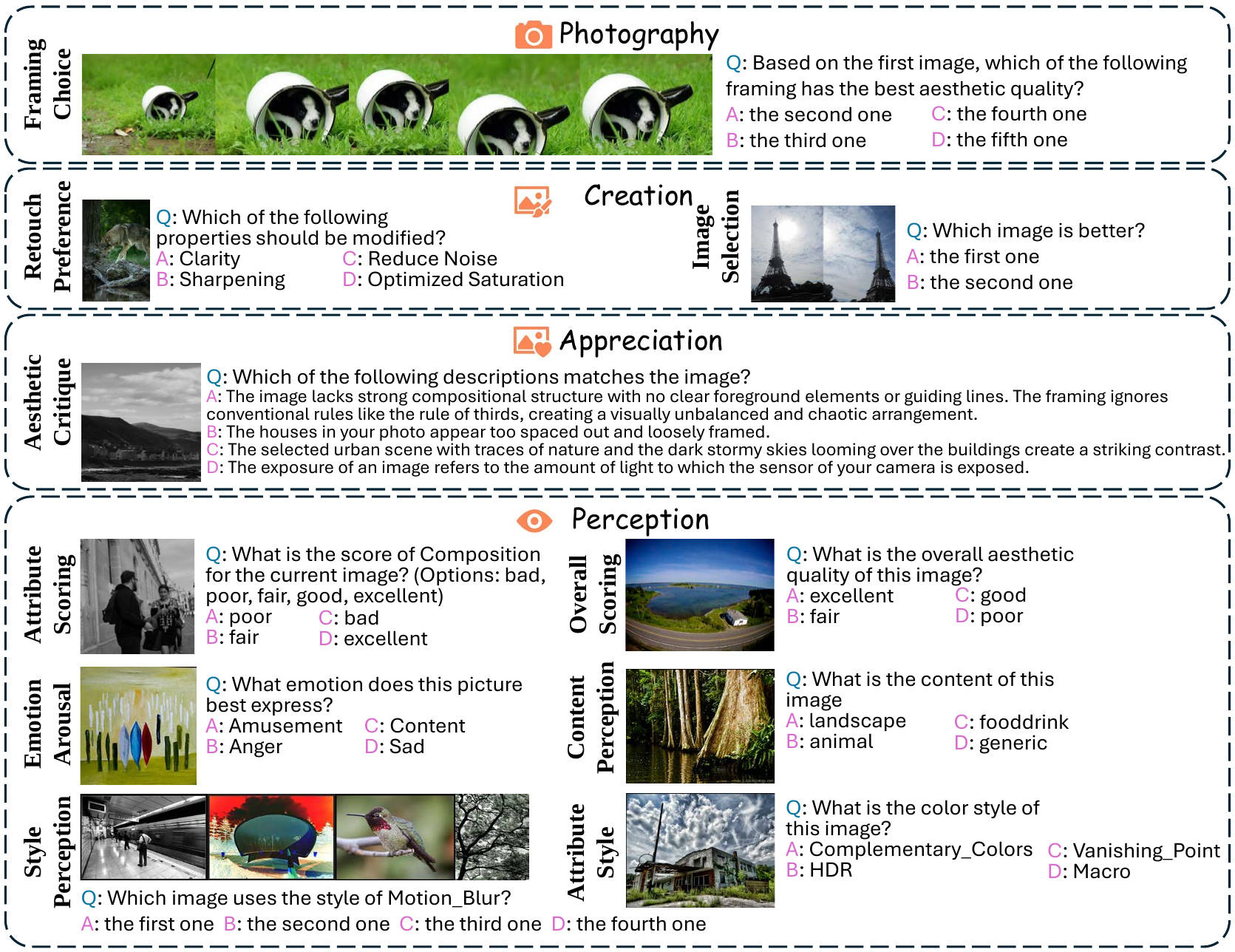} 
	\caption{Examples of each task.}
	\label{fig:task}
\end{figure*}

\section{Related Work}
\noindent\textbf{Image Aesthetic Assessment}. IAA plays a vital role across art, design, and education. Recent deep learning approaches have addressed this task through regression~\cite{murray2012ava,jin2018predicting}, classification~\cite{jin2019aesthetic}, and text generation~\cite{huang2024aesexpert}, often targeting specific attributes. While each task offers practical value, existing research lacks a unified framework, especially for assessing models' aesthetic intelligence. Several benchmark datasets have supported IAA development. Early datasets like PhotoNet~\cite{datta2006studying} and AVA~\cite{murray2012ava} focus on aesthetic scoring, while others incorporate category labels~\cite{luo2011content}, captions~\cite{jin2019aesthetic}, or attribute-level annotations~\cite{he2022rethinking,he2023thinking,yang2022personalized}. Yet, most remain limited to understanding-based evaluations. Crucially, current benchmarks overlook a foundational question: \textit{To what extent has a model truly internalized aesthetic intelligence?} Grounded in the theory of generative learning~\cite{fiorella2016eight}, we argue that IAA should move beyond recognition toward evaluating models' ability to leverage aesthetic knowledge to produce images.

\noindent\textbf{Multi-modal Large Language Models of IAA}. MLLMs have rapidly advanced, beginning with BLIP-2~\cite{li2023blip}, which introduced a Q-Former to bridge visual encoders and LLMs, albeit with limited few-shot capability due to reliance on image-text pairs. Successors such as LLaVA-1.5~\cite{liu2024improved} and LLaVA-NeXT~\cite{li2024llavanext-ablations} have improved alignment and in-context learning. In IAA, models like AesCLIP~\cite{sheng2023aesclip} and VILA~\cite{ke2023vila} leverage contrastive learning or ranking adapters to bridge general vision-language models and aesthetic domains. Zero-shot methods such as CLIP-IQA~\cite{wang2023exploring} and KZIAA~\cite{wang2024keep} exploit vision-language priors for unsupervised evaluation. Instruction-tuned models like AesExpert~\cite{huang2024aesexpert} and UNIAA~\cite{zhou2024uniaa} expand IAA to perception and nuanced critique, while AesBench~\cite{huang2024aesbench} introduces expert-annotated, multidimensional benchmarks to highlight current limitations. Despite these advances, the application of aesthetic knowledge remains underdefined. In this work, we formally introduce the application of aesthetic knowledge within IAA, paving the way for a cognitively grounded, production-oriented aesthetic benchmark.

\section{\ModelName}
\subsection{Task Creation}
To evaluate aesthetic understanding and generation across diverse psychological dimensions, we construct the \ModelName benchmark with a structured taxonomy of ten tasks from eight data sources, as illustrated in Figure~\ref{fig:data}. These tasks span perception, appreciation, creation, and photography, each designed to probe distinct facets of aesthetic intelligence—from perception to production. While all tasks adopt a unified multiple-choice format, they vary substantially in cognitive demands, including interpretive abstraction, visual-semantic alignment, and generative decision-making. This design boosts systematic analysis of MLLMs' capabilities in both recognizing and expressing aesthetic principles, as shown in Table~\ref{tab:benchcompare}.


\noindent\textbf{Overall Scoring \& Attribute Scoring}. The two tasks respectively quantify the overall aesthetic score and the aesthetic scores across distinct attributes. For the latter, based on the three clustered aesthetic attributes, we adopt AMD-A~\cite{jin2023aesthetic} to obtain scores for different attributes. Following~\cite{wu2023q}, we turn numerical values to five-level verbal labels: \textit{excellent, good, fair, poor, and bad}.

\noindent\textbf{Style Perception \& Attribute Style}. Style Perception evaluates a model's capability for aesthetic style recognition using 14 established style tags derived from prior work~\cite{murray2012ava}, while Attribute Style assesses the model's ability to identify stylistic features along defined aesthetic attributes. Through analysis of these style tags and photographer interviews, we classified the styles into three attribute categories: Light, Color, and Composition. Specifically, the Light category includes \textit{Light\_On\_White, Long\_Exposure, Motion\_Blur, Shallow\_DOF, and Soft\_Focus; the Color category comprises Complementary\_Colors, Duotones, HDR, Image\_Grain, and Negative\_Image; and the Composition category contains Macro, Rule\_of\_Thirds, Silhouettes, and Vanishing\_Point}.

\noindent\textbf{Content Perception}. This task is designed to test the capability of a model to understand content, specifically its capacity to accurately identify primary subject matter in images. Following~\cite{murray2012ava}, we obtain nine content tags as the labels: \textit{animal, architecture, cityscape, floral, food drink, generic, landscape, portrait, and still life}.

\noindent\textbf{Emotion Arousal}. This task is designed to evaluate the capacity of a model for emotion congruence with visual content, since aesthetic properties trigger affective responses, while emotional resonance conversely shapes aesthetic judgment ~\cite{joshi2011aesthetics}. Following~\cite{lang1997international}, we obtain the emotion labels of an image.

\noindent\textbf{Aesthetic Critique}. If aesthetic perception constitutes the foundation of aesthetic understanding, then aesthetic criticism serves as its tangible manifestation. Models must perform holistic aesthetic appraisal of images while discerning aesthetic merit. Current approaches exhibit fragmentation~\cite{zhong2023aesthetically}, typically generating image comments first, subsequently identifying aesthetic critiques through lexical analysis, and finally evaluating content. To address this, we employ PCCD~\cite{chang2017aesthetic}, a domain-specific aesthetic critique corpus, for annotation.

\noindent\textbf{Image Selection}. This task marks a pivotal transition from aesthetic evaluation to generative aesthetics. Mirroring photographers' practice of capturing numerous similar scenes for selective editing, preferential image selection constitutes the primary operationalization of aesthetic judgment. Accordingly, we benchmark models' selective editing competence using the Princeton Dataset~\cite{chang2016automatic}, which provides expert annotations for preferential selection tasks among visually comparable images.

\noindent\textbf{Retouch Preference}. This task simulates post-selection retouching workflows in professional photography. We evaluate models' proficiency in selecting optimal enhancement operations, requiring the identification of both most and least appropriate adjustments for given images. Our dataset, curated from Lightroom Gallery~\footnote{https://lightroom.adobe.com/learn/discover}, comprises a comprehensive operation set including color temperature adjustment, exposure correction, and other critical post-processing parameters. Besides, we include cropping evaluation. Specifically, we employ the GAICD dataset~\cite{zeng2020cropping} to formulate benchmark items based on differential aesthetic ratings of multiple crops from identical images. This methodology assesses models' discernment of cropping efficacy through systematic comparative analysis.

\noindent\textbf{Framing choice}. This task evaluates pre-capture aesthetic judgment through post-enhancement analysis. It reflects the ability to distinguish aesthetically optimal viewpoints from suboptimal ones, which is a core skill in photography. Models perform preferential selection among multi-perspective scene captures to identify aesthetically superior viewpoints. Our source data set integrates unused images in AVA and AMD-A datasets.

By organizing these tasks, \ModelName benchmark provides a unified and interpretable framework for evaluating aesthetic intelligence of MLLMs. These tasks are grounded in a psychologically-informed hierarchy that spans perception, appreciation, creation, and photography, enabling fine-grained assessment. Moreover, we design all tasks in a standardized multiple-choice format, ensuring consistency across diverse tasks while allowing flexible extensions in future versions. Examples of each task is shown in Figure~\ref{fig:task}.

\begin{figure}[t] 
\centering
\includegraphics[width=\linewidth]{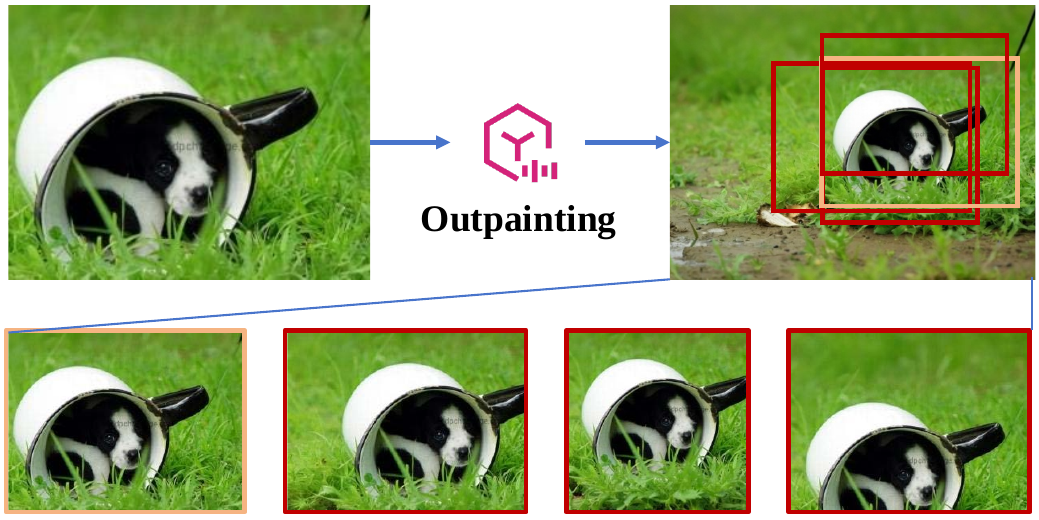}
\caption{An exemplar of aesthetic framing choice Q\&A construction. The orange box marks the pre-expansion framing, and the red box marks the distractor framing.} 
\label{fig:framing}
\end{figure}

\begin{table*}[tb!]
	\centering
            \resizebox{0.98\linewidth}{!}{
			\begin{threeparttable}
				\begin{tabular}{cc|cccccccc|cc}
					\toprule
					\multirow{2}{*}{\textbf{Model}} & \multirow{2}{*}{\textbf{\#Size}} & \multicolumn{2}{c}{\textbf{Perception}} & \multicolumn{2}{c}{\textbf{Appreciation}} & \multicolumn{2}{c}{\textbf{Creation}} & \multicolumn{2}{c}{\textbf{Photography}} & \multicolumn{2}{c}{\textbf{Total}}\\
					\cmidrule{3-12}
					& & Acc & NAcc & Acc & NAcc & Acc & NAcc & Acc & NAcc & Acc & NAcc\\
					\midrule
                    Gemini 2.0 flash~\cite{team2024gemini}$\ddagger$ & -- &0.58&0.43&0.44&0.25&0.48&0.13&0.32&0.10&0.50&0.27\\
                    GPT-4o~\cite{achiam2023gpt}$\ddagger$ & -- & 0.56&0.40&0.57&0.43&0.44&0.08&0.25&0&0.50&0.28\\
                    \midrule
                    CogVLM2-Llama3-Chat~\cite{wang2023cogvlm} & 19.5B & 0.40&0.19&0.34&0.12&0.39&0.01&0.24&-0.02&0.37&0.10\\
                    DeepSeek-VL-Chat-7B~\cite{lu2024deepseek} & 7.3B  & 0.44&0.24&0.35&0.13&0.37&-0.02&0.24&-0.01&0.38&0.11\\
                    InternVL-Chat-1.5~\cite{chen2024far} & 25.5B & 0.44&0.24&0.56&0.42&0.34&-0.07&0.27&0.02&0.42&0.16\\
                    InternVL2-8B~\cite{chen2024far} & 8B & 0.43&0.23&0.55&0.39&0.36&-0.04&0.26&0.05&0.42&0.16\\
                    MiniCPM-V~\cite{yao2024minicpm} & 8B & 0.49&0.30&0.51&0.35&0.39&0.01&0.25&0&0.45&0.20\\
                    Qwen2.5vl~\cite{bai2025qwen2} & 7B & 
                    0.52&0.35&0.58&0.44&0.45&0.09&0.30&0.06&0.50&0.27\\
                    Qwen2.5vl~\cite{bai2025qwen2} & 32B & 0.57&0.42&0.61&0.49&0.46&0.12&0.30&0.07&0.53&0.31\\
                    \midrule
                    AesExpert~\cite{huang2024aesexpert} & 7B & 0.36&0.13&0.36&0.15&0.45&0.12&0.26&0.02&0.39&0.13\\
                    UNIAA~\cite{zhou2024uniaa} & 7B & 0.42&0.22&0.49&0.32&0.36&-0.01&0.25&0&0.41&0.15 \\
                    \midrule
                    amateurs$\dagger$ &  -- &0.48&0.31&0.40&0.20&0.40&0.03&0.35&0.13&0.43&0.18\\
                    professional photographers$\dagger$ & -- &0.51&0.35&0.44&0.25&0.59&0.33&0.83&0.78&0.54&0.35\\
					\bottomrule
				\end{tabular}			
				\end{threeparttable}}
        \caption{Main results across different tasks. $\dagger$ denotes tested on a random subset due to human attention limits. $\ddagger$ if the model consistently fails to generate a valid response after multiple repeated API queries, the instance is marked as incorrect.}
		\label{tab:overall}	
	\end{table*}

    \begin{figure*}[tb!]
		\centering		
		\subfloat[Acc]{
			\includegraphics[width=0.46\linewidth]{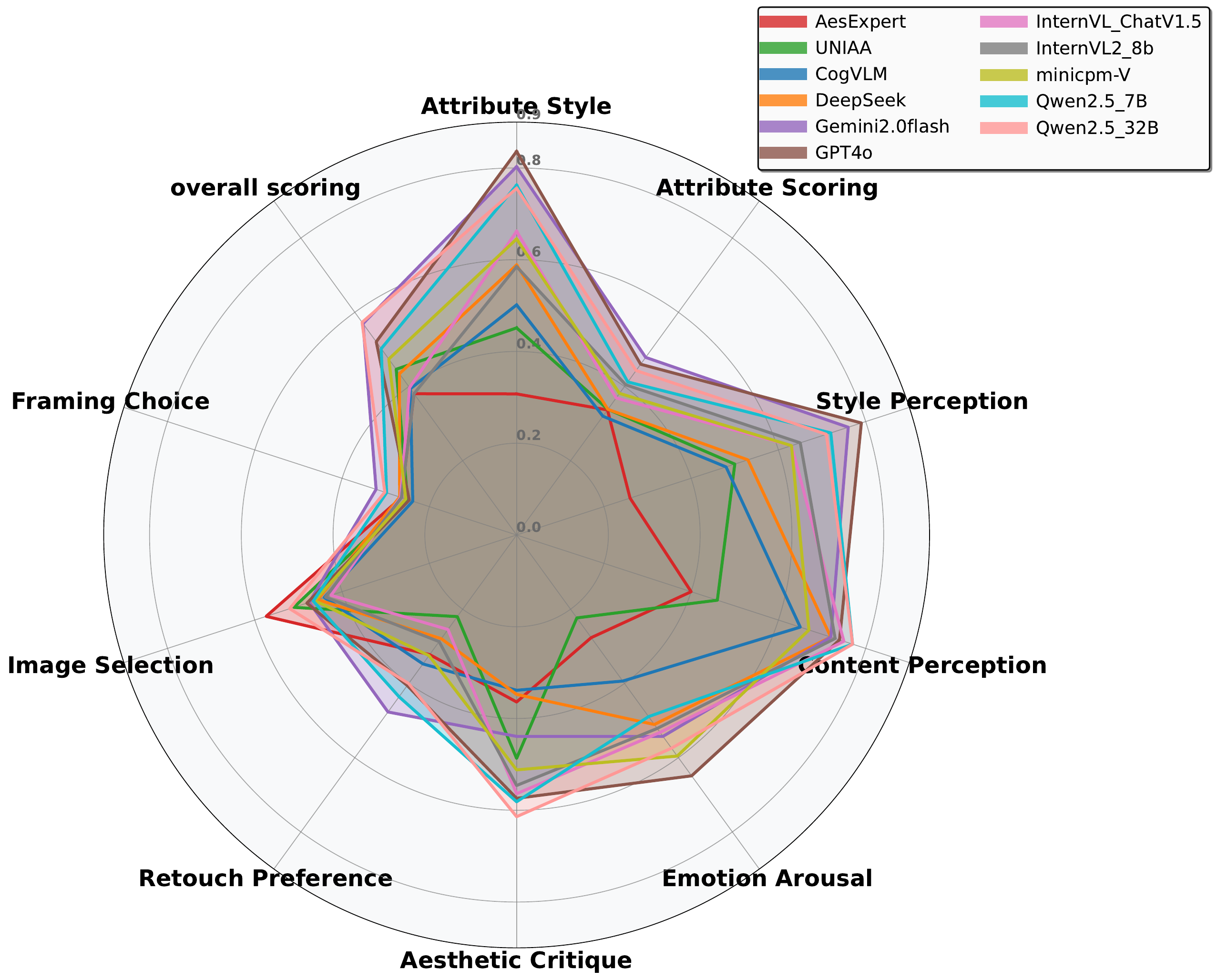}
			\label{fig:accradar}
		}
		\hfill
        \subfloat[NAcc]{
			\includegraphics[width=0.46\linewidth]{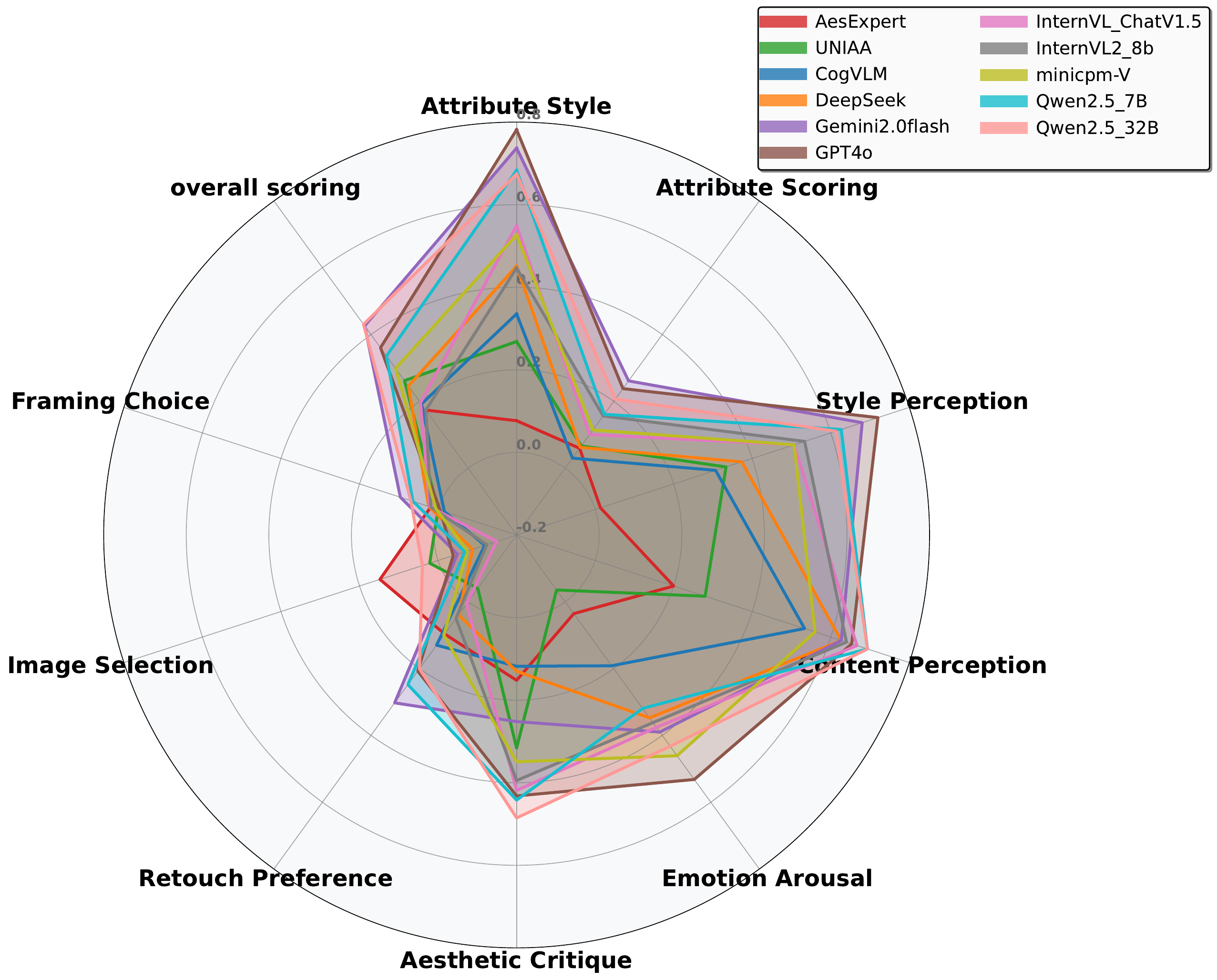}
			\label{fig:naccradar}
		}
		\caption{Sub-task performance of different methods.} 
		\label{fig:radar}
\end{figure*}

    \begin{figure*}[tb!]
		\centering		
        \subfloat[Style]{
			\includegraphics[width=0.24\linewidth]{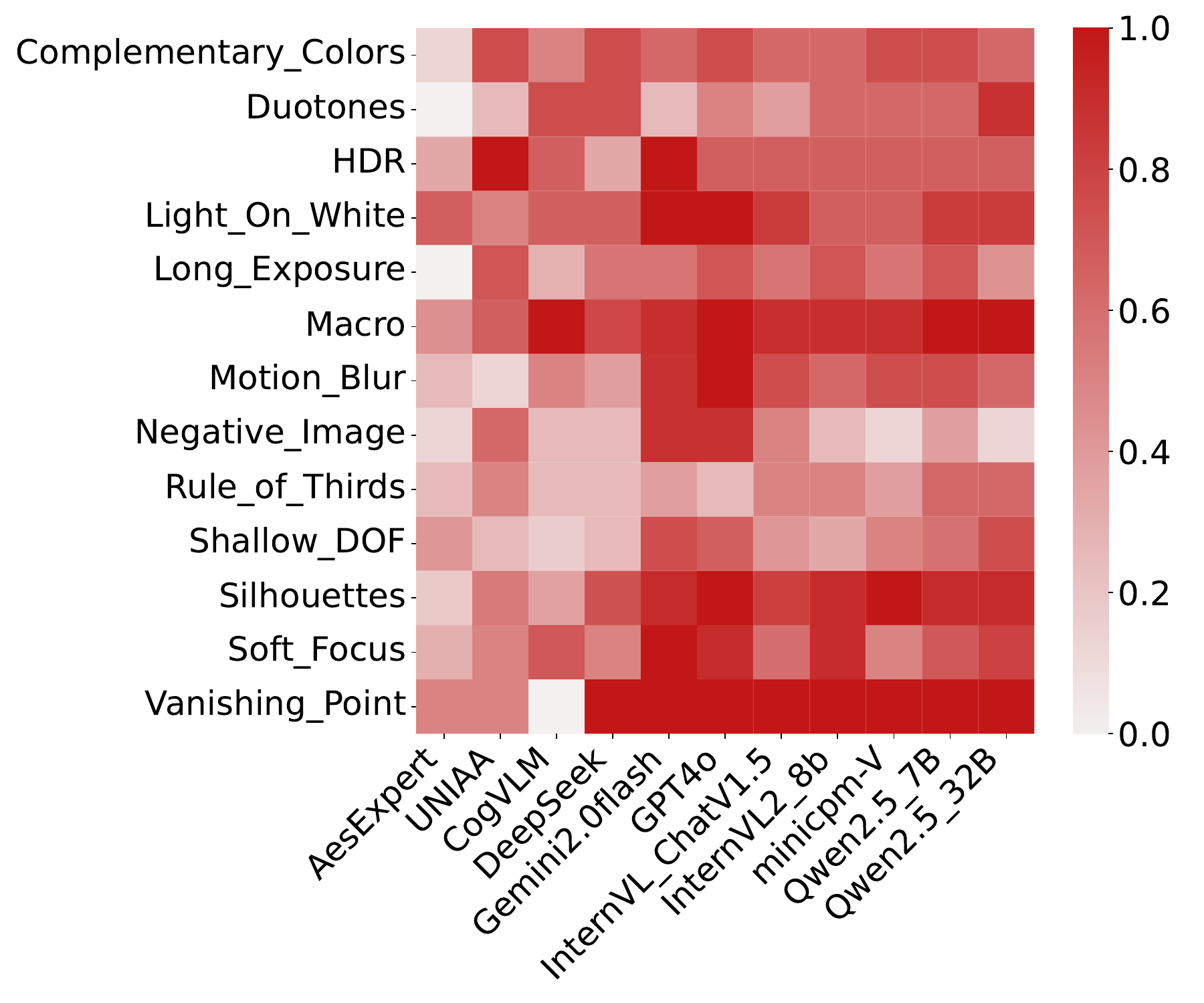}
			\label{fig:styleheat}
		}
        \hfill
        \subfloat[Attribute-based Style]{
			\includegraphics[width=0.25\linewidth]{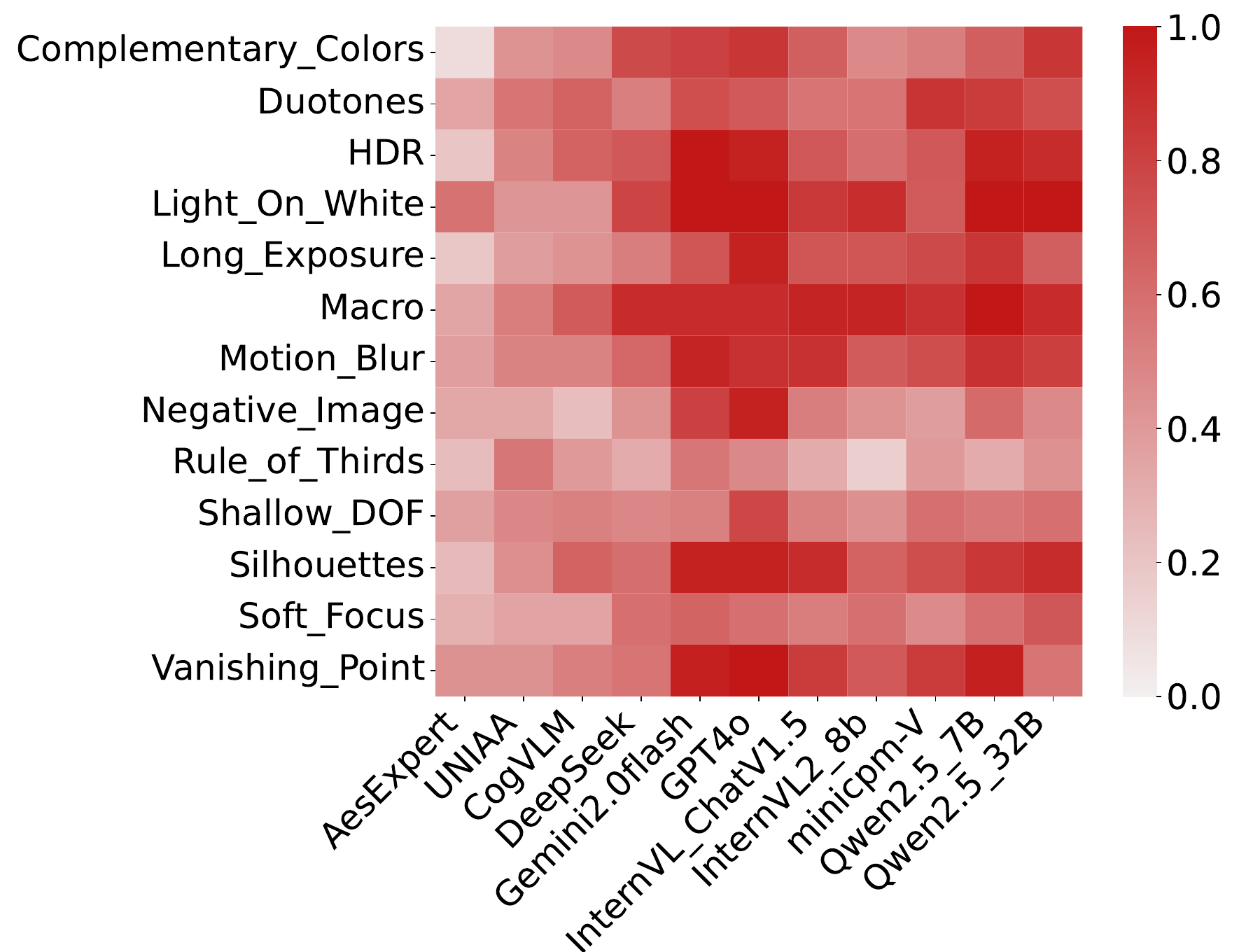}
			\label{fig:attributestyleheat}
		}
        \hfill
        \subfloat[Attribute-based - Raw Style]{
			\includegraphics[width=0.24\linewidth]{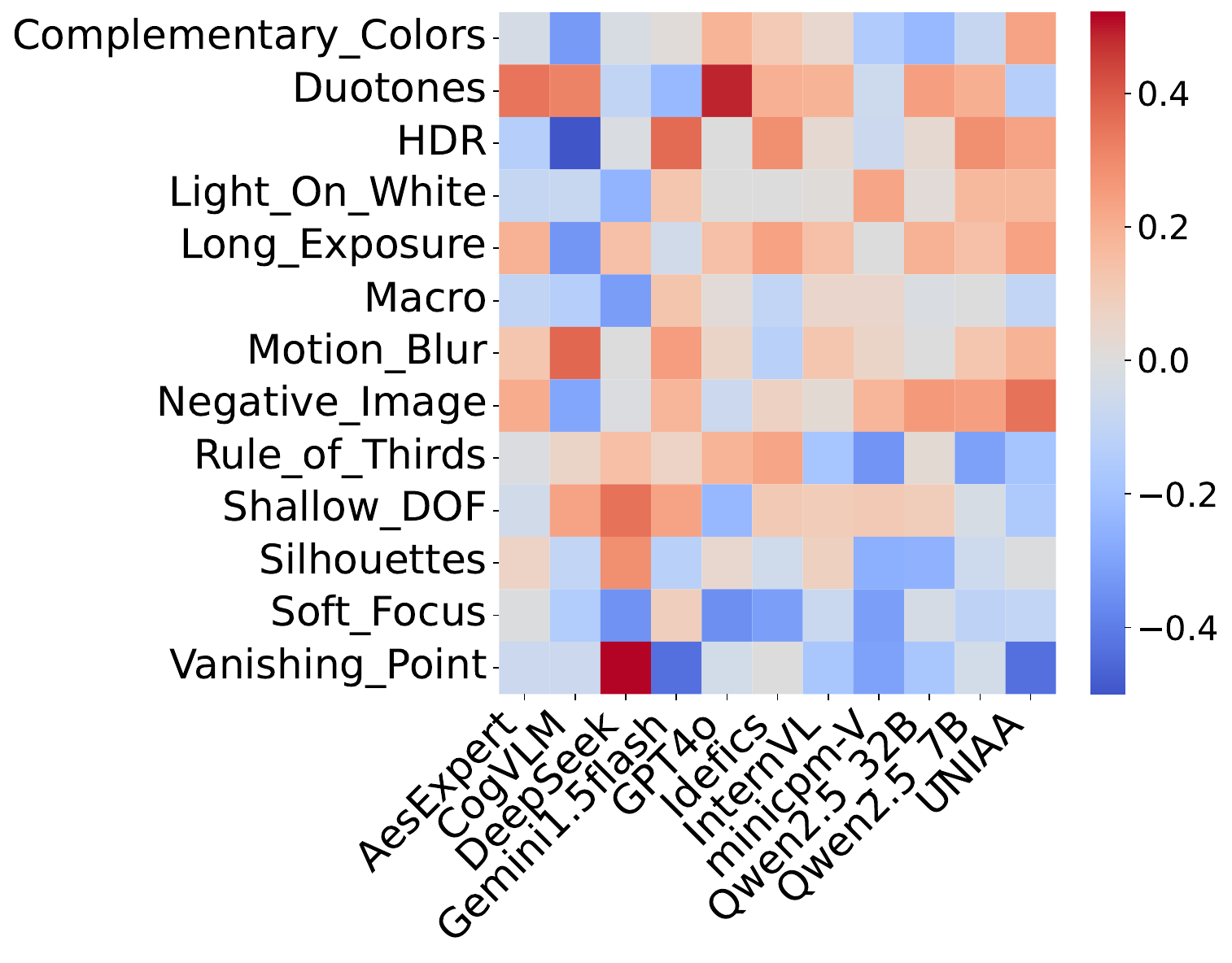}
			\label{fig:stylediffheat}
		}
		\hfill
        \subfloat[Content]{
			\includegraphics[width=0.21\linewidth]{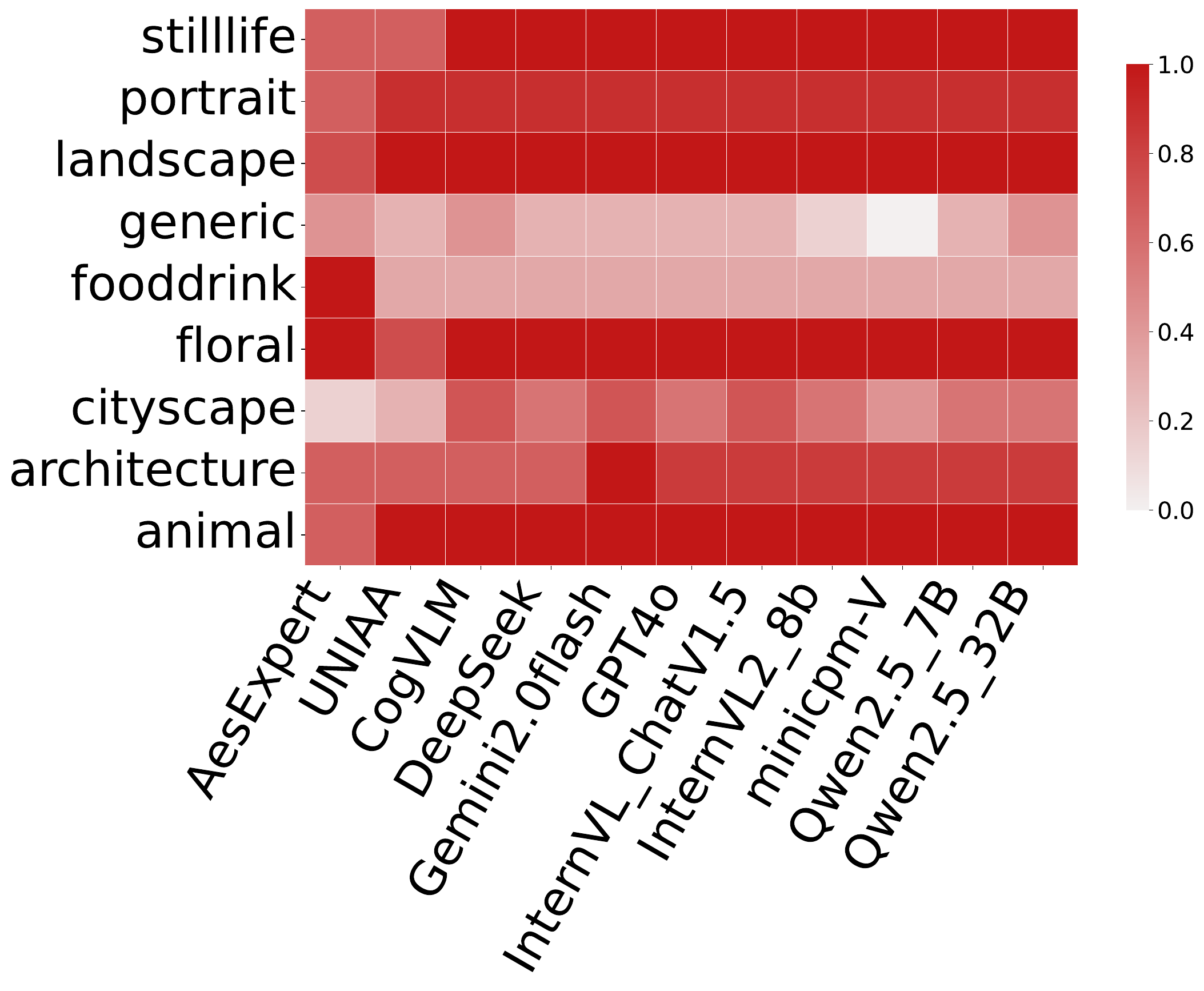}
			\label{fig:contentheat}
		}
		\caption{Fine-grained performance in terms of style and content. (a) Style-wise accuracy of different MLLMs. (b) Style-wise accuracy of different MLLMs given corresponding aesthetic attributes. (c) Difference in style-wise accuracy before and after attribute-based grouping. (d) Content-wise accuracy of different MLLMs.} 
		\label{fig:stylecontentheat}
\end{figure*}

\subsection{Q\&A Construction} 

\noindent\textbf{Overview}. In constructing question-answer pairs for evaluating MLLMs, we standardize multiple-choice question formats to ensure automated and objective assessment. This framework further comprises two distinct settings: (1) single-image setting requiring identification of image characteristics from individual inputs, and (2) multi-image setting involving commonality discovery across image sets and characteristic-to-image matching. Since some MLLMs do not support multi-image input, we concatenate multiple images into a single composite and adjust the corresponding options, following~\cite{meng2024mmiu}. The efficacy of this operation is analyzed in \textit{Appendix A}. We maintain the original aspect ratio of each image during concatenation. This design paradigm effectively evaluates models' bidirectional mapping capabilities between visual content and semantic labels. For each task, we construct the Q\&A as follows and provide detailed prompts in \textit{Appendix B}. 

\noindent\textbf{Construction for Six Sub-Tasks in Perception Task}. This task's question-answer pairs leverage existing publicly available datasets with high-quality annotations. Regardless of whether inputs consist of single images or tiled multi-image compositions, correct answers correspond to ground-truth labels, while distractors are randomly sampled from incorrect label pools.

\noindent\textbf{Construction for Aesthetic Critique}. To ensure quantifiable task design, we utilize textual annotations from the PCCD dataset, where each image has comments covering multiple attributes. We select the critiques for a random attribute as the correct answer, transform the remaining attribute critiques into contradictory descriptions using DeepSeek, and randomly sample from them as distractors.

\noindent\textbf{Construction for Image Selection}. The QA construction in this task follows the paradigm used in the Princeton Dataset, where a question is formed by selecting a single image from an image series. The correct answer corresponds to the selected image, while the distractors are drawn from the remaining images in the same series.

\noindent\textbf{Construction for Retouch Preference}. For Lightroom, we curate the top ten edits per image by different professionals, ranked by aesthetic scores. To reduce stylistic bias, we extract operations consistently applied or avoided across photographers and evaluate whether the model distinguishes essential from undesirable actions. For cropping, the model selects the most aesthetic option from four candidate crops.  

\noindent\textbf{Construction for Framing Choice}. For our Framing Choice task, we simulate a realistic shooting scenario in which a photographer views a panoramic scene and selects the best framing for capture. To implement this, we take high-scoring images from unused images in existing datasets and apply outpainting to obtain a panorama that represents the scene. The original high-quality image used for expansion serves as the photographer’s chosen framing. We then create distractors by cropping adjacent regions of the outpainted panorama at 1:1, 4:3, 16:9, and the original image's aspect ratio. To mitigate artifacts introduced by outpainting, we also extract the original image region from the outpainted panorama at the corresponding coordinates. The whole process is shown in Figure~\ref{fig:framing}.

\subsection{Evaluation Metrics}
\noindent\textbf{Accuracy} (Acc) is the proportion of questions for which the model selects the correct answer.

\noindent\textbf{Chance-normalized Accuracy} (NAcc) accounts for varying levels of random guessing. Specifically, for a task with $n$ answer options, the normalized score is computed as:
\begin{equation}
\label{eqn:nacc}
\mathrm{NAcc} = ( \mathrm{Accuracy} - 1/n ) / (1 - 1/n)
\end{equation}
\section{Experiments}
\subsection{Baselines}
\noindent\textbf{Models}. In this paper, we evaluate three mainstream approaches on \ModelName. The first category is instruction-tuned IAA MLLMs, like AesExpert~\cite{huang2024aesexpert}. The second category is closed-source MLLMs, like GPT-4o. The third category is open-source MLLMs, such as Qwen-VL. For all models, we employ a standardized prompt to ensure consistency across evaluations.

\noindent\textbf{Human}. We recruit ten amateurs and five professional photographers to participate in the evaluation. To mitigate aesthetic fatigue caused by prolonged testing, we randomly sample a subset of 100 questions from the full benchmark and remove obviously invalid responses (See \textit{Appendix C}).

\subsection{Results}
\noindent\textbf{Main Results}. The overall results are shown in Table~\ref{tab:overall} and Figure~\ref{fig:radar}. The main observations are summarized as follows:

(1) \textbf{General vs. Aesthetic-Specific Models}. General MLLMs perform better overall; Qwen2.5-VL-32B, GPT-4o, and Gemini-flash lead and even surpass aesthetic-specific models. Aesthetic-specific models are mid-tier overall but show an advantage on Creation task.

(2) \textbf{Top MLLMs vs. Human}. Creation and Photography yield the largest model–photographer separation; for example, on Photography professionals reach 0.83/0.78 Acc/NAcc, whereas the best model attains only 0.46/0.12. Besides, amateurs outperform top MLLMs on the Photography task, though they do not excel overall. On Perception and Appreciation, the gap between professionals and top models is small, aligning with current capabilities of MLLMs and supporting the validity of our benchmark.  

(3) \textbf{Amateurs vs. Professional Photographers}. Compared with professional photographers, amateurs exhibit a clear gap, especially on Creation and Photography. This finding validates the discriminative power of our benchmark.

(4) \textbf{Scale Effects}. Scale effects are clear on Perception and Appreciation but weak on Creation and Photography. Qwen2.5-VL-32B improves over the 7B model by about 10\% and 5\% on Perception and Appreciation, respectively, but shows only marginal gains on Creation and Photography.

(5) \textbf{Task Difference}. Task difficulty for models increases from Perception to Appreciation to Creation to Photography. From the sub-task perspective, Content Perception scores for most models approach the outer bound, whereas Framing Choice remains consistently low across models. This pattern aligns with current MLLM capabilities. These phenomena are not captured by existing benchmarks, indicating that our benchmark remedies their limitations.

\noindent\textbf{Style-wise Results}. We analyze model performance across different styles, which reflects models' ability to recognize higher-order aesthetic patterns beyond surface features. Figure~\ref{fig:stylecontentheat}\subref{fig:styleheat} shows results of \textit{Style Perception}. Figure~\ref{fig:stylecontentheat}\subref{fig:attributestyleheat} shows the results of style classification given corresponding aesthetic attributes. Figure~\ref{fig:stylecontentheat}~\subref{fig:stylediffheat} presents their accuracy differences. The results show that MLLMs performs well on composition-related styles (e.g., \textit{Vanishing\_Point}). After providing the aesthetic attributes associated with each style, most models improve. This aligns with instruction following in large models and supports the validity of our benchmark. The largest gains appear for Long\_Exposure, Motion\_Blur, and HDR, which require integrating multiple capture and post-processing techniques; attribute cues reduce the reasoning burden. Unnatural or strongly stylized effects remain difficult but still improve; for example, Negative\_Image and Duotones are weak in Figure~\ref{fig:stylecontentheat}\subref{fig:styleheat} and rise in Figure~\ref{fig:stylecontentheat}\subref{fig:attributestyleheat} yet still lag others, suggesting limited priors for artificial renderings. Overall, attribute cues lift weaker models while largely preserving the ranking of stronger ones, indicating that this sub-task increases the overall signal-to-noise ratio without compromising comparability.

\noindent\textbf{Content-wise Results}. We analyze model performance across semantic content categories (e.g., portrait), as aesthetic judgment often varies with content. As shown in Figure~\ref{fig:stylecontentheat}~\subref{fig:contentheat}, most categories score near the ceiling, indicating reduced content bias in our \ModelName as limited dominance of cues like portrait or animal. Notable drops for generic and fooddrink reveal weaknesses in low-structure scenes, demonstrating fine-grained diagnostic capability of our \ModelName. Within model families, scores rise with size while profiles stay stable, making this dimension repeatable and scalable. 
\begin{figure}[t] 
\centering
\includegraphics[width=\linewidth]{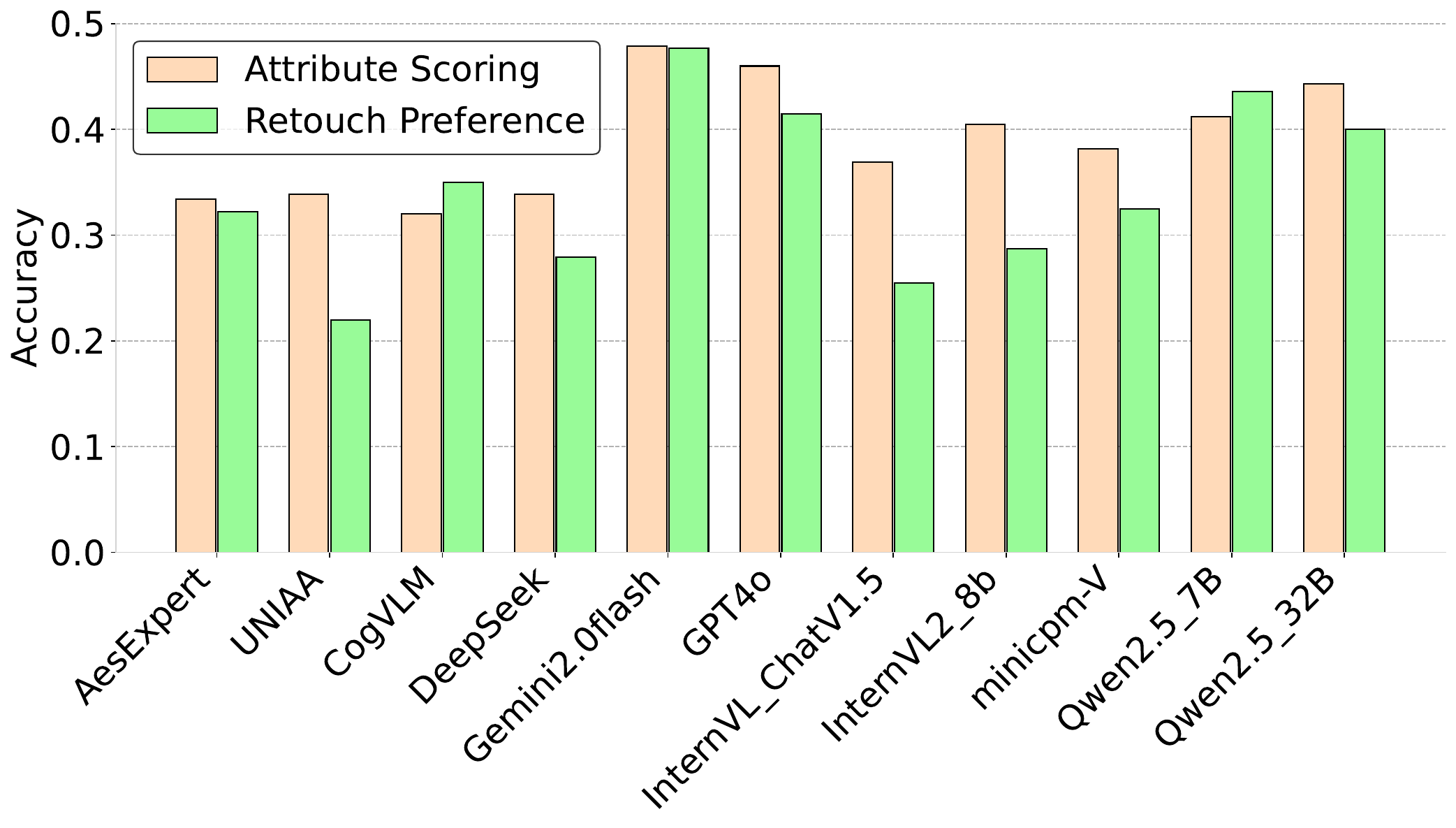}
\caption{Performance on \textit{Attribute Scoring} and \textit{Retouch Preference} sub-tasks.} 
\label{fig:attributecompare}
\end{figure}
\begin{figure*}[tb!]
	\centering
	\includegraphics[width=0.96\linewidth]{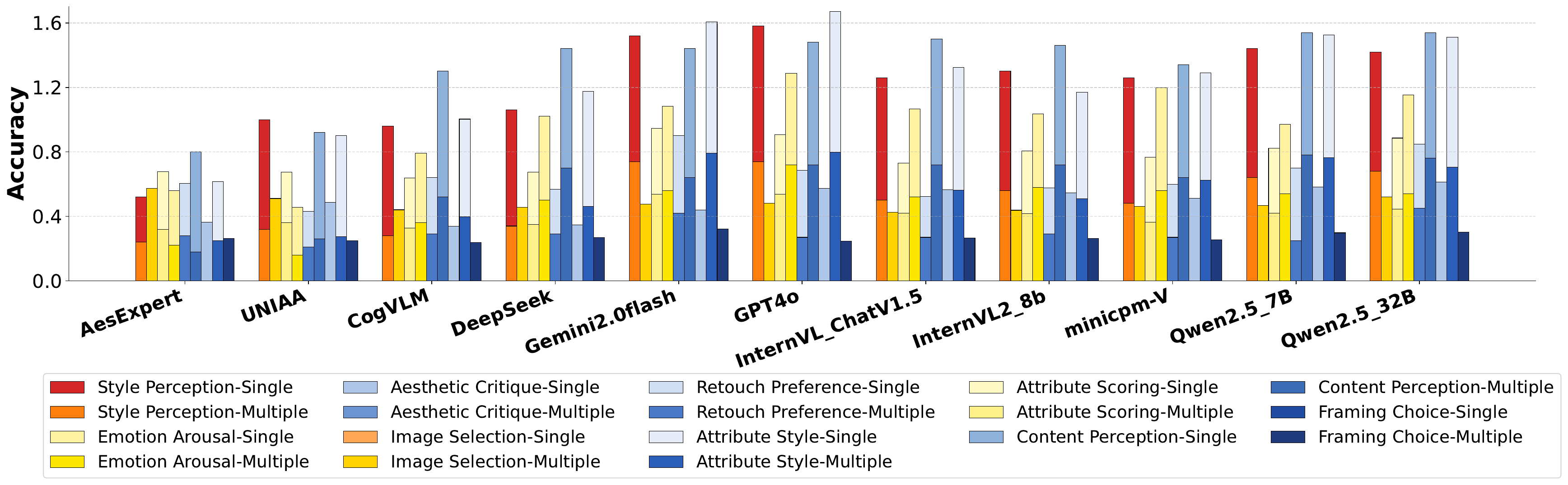} 
	\caption{Accuracy of each task with single- and multi-stimulus inputs. Bars are stacked: lower and upper segments indicate multi-stimulus and single-stimulus accuracy, respectively. \textit{Framing Choice} and \textit{Image Selection} are assessed only with multi-stimulus inputs; \textit{Aesthetic Critique} only with single-stimulus inputs.}
	\label{fig:singlemultiple}
\end{figure*}

\noindent\textbf{Attribute-wise Results}. In our \ModelName, we include two tasks that assess models from the perspective of aesthetic attributes: \textit{Attribute Scoring} and \textit{Retouch Preference}. We separate perceiving attributes from applying them in edits. The results in Figure~\ref{fig:attributecompare} reveal a consistent gap between understanding and production. The results demonstrate discriminative power of our \ModelName, matching our design objective. 

\noindent\textbf{Format-wise Results}. Our \ModelName includes two input formats: single-image and multi-image prompts. We conduct a systematic analysis to examine how question format influences model performance across aesthetic tasks. Our \ModelName includes two input formats: single-image and multi-image prompts. We conduct a systematic analysis to examine how question format influences model performance across aesthetic tasks. Figure~\ref{fig:singlemultiple} shows the accuracy of each task under single-image (upper segment) and multi-image (lower segment) stimulus. The results suggest taht models with overall strong performance (e.g., GPT4o) are relatively robust across both single-image and multi-image formats, indicating a more stable and adaptable aesthetic intelligence. By designing tasks with both single-image and multi-image prompts, \ModelName functions as a format-aware stress test that diagnoses robustness and generalization. Consistent performance across formats signals format-invariant aesthetic competence, while divergences reveal annotation-fitting and transfer limitations.

\begin{figure}[t] 
\centering
\includegraphics[width=\linewidth]{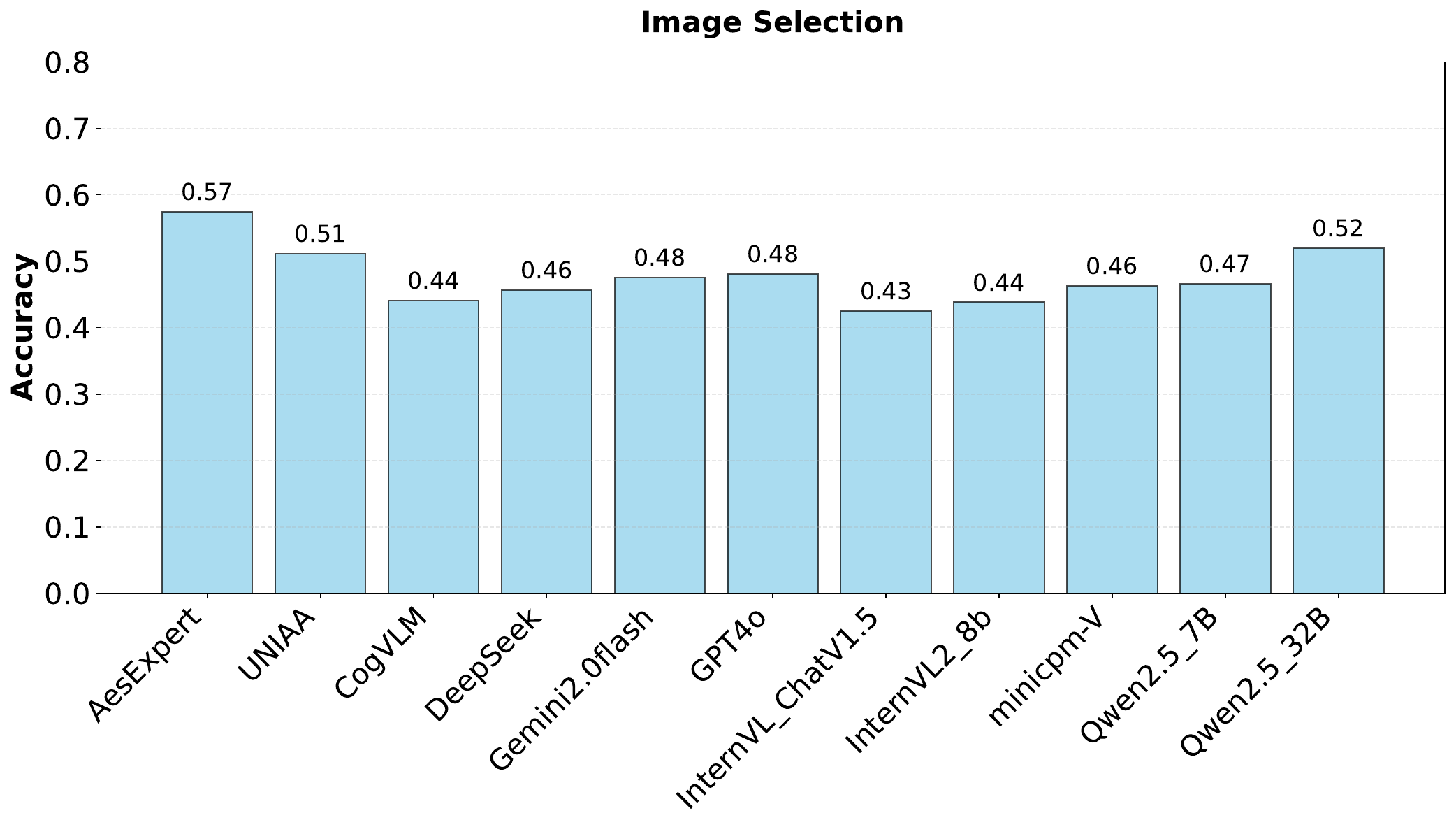}
\caption{A comparative analysis between AesExpert and other open-source MLLMs on single-stimulus questions in \textit{Attribute Scoring}.} 
\label{fig:scorecompare}
\end{figure}

\subsection{Analysis}
\noindent\textbf{MLLMs require targeted fine-tuning for aesthetics}. Simply increasing model size without aesthetic instruction does not necessarily lead to improved aesthetic performance. For example, while Qwen2.5-VL-32B shows modest gains over Qwen2.5-VL-7B on perception and appreciation tasks, its performance on creation and photography tasks actually remains largely unchanged, highlighting the limitations of scaling alone in capturing higher-order aesthetic intelligence (Table~\ref{tab:overall}).

\noindent\textbf{Enhancing the aesthetic capabilities of MLLMs requires more than fitting to instruction-tuning data}. Besides, MLLMs must learn to align abstract aesthetic concepts with corresponding visual features. Simply increasing model size without aesthetic instruction does not necessarily lead to improved aesthetic performance. The generally decreased performance on multi-stimulus tasks further suggests that most models struggle to establish a clear bidirectional mapping between aesthetic concepts and visual features (Figure~\ref{fig:singlemultiple}). In other words, they may recognize that a given image statistically correlates with a particular concept, but lack an understanding of how that concept is constructed through concrete visual elements. 

\noindent\textbf{Instructing MLLMs through preference learning demonstrably improves their understanding of aesthetic preferences}. Although models explicitly fine-tuned for aesthetics, such as AesExpert, do not consistently outperform others across all tasks, they show a clear advantage in Image Selection sub-task. This indicates that aesthetically finetuned MLLMs require more comprehensive benchmarks to fully demonstrate their effectiveness.
 (Figure~\ref{fig:scorecompare}). 

\section{Conclusion}
We present \ModelName, a unified benchmark designed to evaluate the aesthetic understanding and production capabilities of MLLMs. \ModelName spans ten tasks across perception, appreciation, creation, and photography, each framed as a multiple-choice question to ensure consistency and scalability. Our benchmark highlights core challenges in aesthetic intelligence, such as attribute-level interpretation, emotional alignment, and compositional judgment. Empirical results reveal that while MLLMs show promising perceptual skills, their performance on production remains limited. We hope \ModelName will provide a foundation for advancing aesthetic intelligence in multimodal systems and inspire future work in cognitively grounded vision-language learning.

\bibliography{custom}




\end{document}